# Text Understanding and Generation Using Transformer Models for Intelligent E-commerce Recommendations

Yafei Xiang[1]*, Hanyi Yu[2], Yulu Gong[3], Shuning Huo[4], Mengran Zhu[5]

[1]Computer Science, Northeastern University, Boston, MA, USA, xiang.yaf@northeastern.edu
[2]Computer Science, University of Southern California, Los Angeles, CA, USA, hanyiyu@usc.edu
[3]Computer & Information Technology, Northern Arizona University, Flagstaff, AZ, USA
yg486@nau.edu
[4]Statistics, Virginia Tech University, Blacksburg, VA, USA, shuni93@vt.edu
[5]Computer Engineering, Miami University, Oxford, OH USA, mengran zhu0504@gmail.com
*Corresponding author: [Yafei Xiang E-mail: xiang.yaf@northeastern.edu]

**ABSTRACT:** With the rapid development of artificial intelligence technology, Transformer structural pre-training model has become an important tool for large language model (LLM) tasks. In the field of e-commerce, these models are especially widely used, from text understanding to generating recommendation systems, which provide powerful technical support for improving user experience and optimizing service processes. This paper reviews the core application scenarios of Transformer pre-training model in e-commerce text understanding and recommendation generation, including but not limited to automatic generation of product descriptions, sentiment analysis of user comments, construction of personalized recommendation system and automated processing of customer service conversations. Through a detailed analysis of the model's working principle, implementation process, and application effects in specific cases, this paper emphasizes the unique advantages of pre-trained models in understanding complex user intentions and improving the quality of recommendations. In addition, the challenges and improvement directions for the future are also discussed, such as how to further improve the generalization ability of the model, the ability to handle large-scale data sets, and technical strategies to protect user privacy. Ultimately, the paper points out that the application of Transformer structural pre-training models in e-commerce has not only driven technological innovation, but also brought substantial benefits to merchants and consumers, and looking forward, these models will continue to play a key role in e-commerce and beyond.

**Keyword list:** Large language model; E-commerce intelligent recommendation; Transformer pre-training model; Text understanding and generation

## 1. INTRODUCTION

E-commerce Recommendation System (RecSys) is a personalized system based on users' historical behavior and interest preferences, which can provide users with accurate and personalized product recommendations and promote users' shopping experience and consumption satisfaction. A mature RecSys usually adopts the cascade structure of the pipeline, including various modules such as recall, rough discharge, fine discharge, and rearrangement, which has a high degree of professional field characteristics. With the explosion of ChatGpt, large language models (LLMS) began to emerge in more and more fields. LLM is a natural language processing technology based on deep learning, which is able to learn the rules and patterns of language from large-scale corpora. When the model parameters break through a certain scale, the performance is significantly improved, and LLM begins to show emergence ability and generalization ability. A large number of general world knowledge is stored in the huge parameters, and it has language understanding and expression ability. Compared with LLM, RecSys is a data-driven system that relies on e-commerce ID system to model users or items, lacks semantic and external knowledge information, and has problems such as information cocoon, cold start, lack of diversity, and inability to cross-domain recommendation. However, the latter lacks the proprietary data information in the recommendation field, does not have the sequence processing and memory ability of the traditional recommendation model, and has high computational complexity and high training and reasoning cost.

The Transformer model has significant advantages in intelligent recommendation of large language models. Its self-attention mechanism allows it to better understand the relationships between text sequences and thus more precisely capture the user's preferences and intentions. After pre-training large-scale corpus, Transformer model has rich language knowledge and can generate more smooth and natural recommendation text. In addition, the Transformer model can also personalize recommendations and generate customized recommendations based on the user's historical behavior and preferences, improving the accuracy of the recommendation system and user satisfaction.

Therefore, this paper focuses on text understanding and generation in e-commerce intelligent recommendation based on Transformer pre-training model in the field of large language model, analyzes the current development status of large language model, and discusses the application advantages of BERT, Transformer and other models. Finally, the application of intelligent recommendation of e-commerce products is discussed through practical experiments.

## 2. THEORETICAL OVERVIEW

### 2.1 Transformer model

Transformer is a seq2seq model proposed by Google in 2017 and applied to neural machine translation. Its structure completely uses self-attention mechanism to complete global dependency modeling of source language sequence and target language sequence. Transformer consists of encoders and decoders. Figure 1 shows this structure, with Encoder and Decoder structures on the left and right, respectively, each consisting of several basic Transformer Encoder/Decoder blocks (N × represents N stacks).

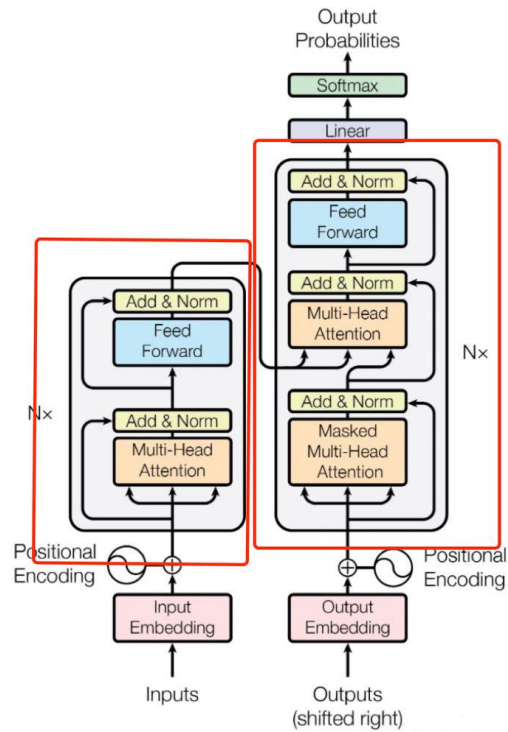

**Figure 1:** Transformer model architecture diagram

(1) Encoder and decoder layer
Transformer's architecture is divided into encoders and decoders.
Transformer's encoder is stacked with N identical layers, each with two sub-layers. The first sub-layer is a multi-head self-attention mechanism, and the second sub-layer is a simple position fully connected feedforward network. There is a residual connection around each subelayer, followed by layer normalization. Residual connections help avoid the problem of disappearing gradients in deep networks. Layer normalization standardizes the output of each sub-layer and helps stabilize the training process.
The decoder part is also made up of N identical layers. In addition to the two sublayer in each layer (multi-head self-attention layer and feedforward network), each layer of the decoder also contains a third sublayer that performs multi-head attention operations on the output of the encoder.
For an Input text sequence, each word is first converted to its corresponding vector representation by the Input Embedding layer. Usually a vector representation is created directly for each word. Since there is no information in the Transfomer that represents the relative position relationships between words, Positional Encoding is added to the word embeddings. Specifically, the position of each word in the sequence corresponds to a vector. This vector is then added to the word representation and sent to a subsequent module for further processing. During the training process, the model will automatically learn how to use this part of the location information.
（2）Multi-Head-self-Attention

The core of Transformer is its self-attention mechanism. It allows the model to consider all other words in the sentence while processing each word, effectively capturing long-distance dependencies. This is achieved by calculating the relationship between the query (Q), key (K), and value (V), where the attention score is calculated by the following formula:

$$\text{Attention}(Q, K, V) = \text{softmax}(\frac{QK_T}{d_k})V \quad (1)$$

Transformer uses a multi-head attention mechanism that splits attention across different presentation subspaces, allowing the model to understand information from multiple perspectives at the same time. This mechanism is implemented in the following ways:

$$\text{MultiHead}(Q, K, V) = \text{Concat}(\text{head}_1, ..., \text{head}_h)W^O \quad (2)$$

For Multi-Head Attention, Scaled Dot-Product Attention and Multi-Head Attention are used. Scaled dot-product Attention means that the attention of the Dot Product is added to the additional scale (self-attention divided by the square root of key), in practice, query, key, and value are all merged into the matrix Q, K, and V operations:

Multi-Head Attention refers to the linear mapping of Q, K and V above to generate multiple heads. Each head is mapped to a different space, and the focus learned is likely to be different, so as to enrich the information brought by the model.

(3) Features and advantages of Transformer

Parallelization capability: Due to its non-recursive nature, Transformer enables efficient parallel processing when processing serial data.

Long Distance Dependency Handling: The self-attention mechanism enables Transformer to efficiently handle long distance dependencies, addressing the limitations of traditional RNNS and LSTMS in this regard.

Flexibility and versatility: Transformer is suitable for a variety of different tasks, from text translation to content generation, showing great flexibility and wide applicability.

Transformer's design of large language model processing is a major breakthrough, but also provides a new perspective on the future development of machine learning and artificial intelligence. Its powerful performance and wide application potential make it one of the key factors for current and future technological advances. Therefore, by combining the text processing and understanding generation functions of Transformer model and e-commerce intelligent recommendation system, this paper realizes the practical application and advantages of large language model.

**2.2 Application of Transformer model in E-business recommendation**

(1) LLM and Intelligent recommendation

In the traditional recommendation field, commodity and text information are often represented by numerical ID without semantic meaning, and the sparse one-hot ID feature coding is designed as a simple Embedding Look-up Table. Even features rich in text semantics (such as the title of a product, attributes) are unified into ID codes, resulting in semantic information loss. With the rise of language models, an intuitive approach is to use language models as embedded representations of text information such as product titles/attributes obtained by encoders, combined with the ID-based one-hot encoding mode of recommendation systems. Typical work includes: U-BERT and Transformer[8] encoding the user comment content to enhance the user's personalized vector representation, and eventually obtain a dense embedding vector; UniSRec[9] achieves the goal of cross-domain sequence recommendation by encoding the item title/ user behavior sequence.

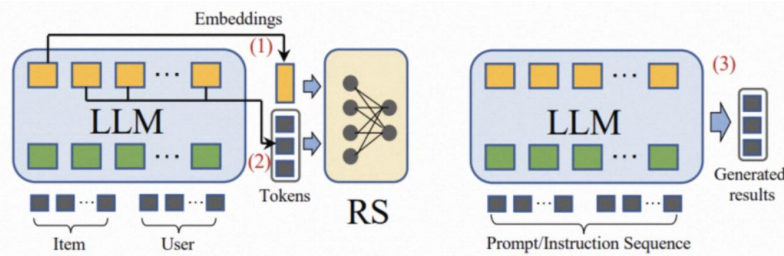

**Figure 2:**Two fusion modes of LLM and recommendation [1] : 1. LLM+ Recommendation (a) 2. LLM as recommendation (b)

In the field of e-commerce intelligent recommendation, LLM is used to summarize the original corpus information of goods/users and generate a concise semantic expression for subsequent recommendation modules. For example, Liu et al. [9] proposed a generative news recommendation framework based on LLM. By building data such as news headlines, categories, etc. into prompt prompts, the LLM is motivated to generate relevant information such as news summaries, personalized news, etc., based on its general knowledge. On the one hand, the generated information is used to iteratively optimize LLM generation; On the other hand, it is used to train the news recommendation model and supplement the knowledge information of the recommendation model.

(2) Data processing:

When preparing and preprocessing e-commerce text data, we take the following steps:

1. Data cleaning and standardization: The original text data is cleaned and standardized, including the removal of special characters, punctuation, stop words, etc., and the text is converted to a unified format.

2. Word Embedding: Converting text data into a dense vector representation, we use a pre-trained word embedding model such as Word2Vec, GloVe or BERT to obtain the semantic information of the word.

3. Sequence filling and truncation: Since the Transformer model requires the same length of input sequences, we fill or truncate the sequences to make them have the same length.

4. Data Augmentation: In order to increase the diversity of data, we employ data augmentation techniques such as random deletion, replacement or insertion of words, and text translation.

(3) Recommendation strategy:

Our recommendation strategy is mainly based on the following aspects:

1. User behavior modeling: We use Transformer model to model user history behavior sequence, including browsing history, purchase history, etc., to capture user interests and preferences.

2. Product Description Modeling: Similarly, we use the Transformer model to model the product description text to understand the features and attributes of the product.

3. Personalized recommendation generation: Combining user behavior and product description representation, we use Transformer model to generate personalized recommendation results. This involves using the encoder to generate user representations and product representations, and then using the decoder to generate a list of recommendations.

4. Multi-task learning: In order to further improve the recommendation effect, we can optimize tasks such as click rate prediction and conversion rate prediction as multiple goals, and use mechanisms such as MMoE to carry out multi-task learning.

## 3.  METHODOLOGY

In the experiment part of this paper, Taobao, one of the largest online e-commerce platforms in China, is the driving force for its continuous transactions through the accurate product recommendation algorithm. The user has clicked or purchased the product in the history, and then recommends the product that the user is interested in again based on the historical behavior data of the user, in which the user behavior sequence data plays an important role. Considering the advantages of Transformer model in the field of NLP in sequence modeling, Transformer unit is introduced into the recommendation algorithm model of Taobao to learn the sequence information of users' historical behaviors.

### 3.1 Transformer model data for user behavior

In the rank stage, we model the recommendation task as a click-through rate (CTR) prediction problem, defined as follows: The sequence of behavior of a given user $S(u) = \{v_1, v_2, ... v_n\}$ is clicked by user u, and we need to learn a function F to predict the probability of a user clicking $v_t$, where $v_t$ is one of the candidate items. Other features include user profile, context, item, and cross features.
We build BST on top of WDL, and the overall architecture is shown in Figure 3. From Figure 3, we can see that it follows the popular Embedding&MLP paradigm, where the previous click item and associated features are first embedded in a low-dimensional vector and then input into the MLP. The key difference between BST and WDL is that we've added a Transformer layer that learns to better represent items that users click on by capturing the underlying sequential signal. In the following sections, we cover the key components of the BST from the bottom up: the Embedding layer, the Transformer layer, and the MLP.

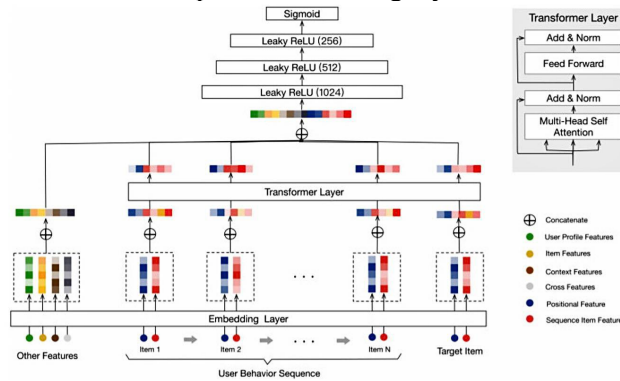

**Figure 3:** E-commerce intelligence recommends the overall architecture of BST

BST takes the user's sequence of actions as input, including the target item and other characteristics. It first embeds these input features as low-dimensional vectors. To better capture the relationships between items in a behavioral sequence, the Transformer layer is used to learn a deeper representation of each item in the sequence. The three-layer MLP is then used to learn the hidden feature interactions by connecting the embed of other features to the output of the Transformer layer, and the Sigmoid function is used to generate the final output.
Note: "Location features" are subsumed into "sequence features".

### 3.2 Embedding Layer

As the standard configuration of the recommended algorithm model of deep learning, the original dense and sparse features are transformed into the embedding vector features, which are mainly divided into two types of features:

Other Features: Here, user profile features, item features, context features, cross features, etc. are collectively referred to as other features, specifically as shown in the following figure. This is mainly because the BST model mainly studies the processing of sequence features, so other features other than sequence features are uniformly processed here.

Sequence Features: The behavioral Sequence Features here mainly include two types of Features, one is Sequence Item Features and the other is Positional Features.

Table 1:Intelligent recommendation sequence data

| User | Item | Context | Cross |
|---|---|---|---|
| gender | category_id | math_type | age*item_id |
| age | shop_id | disply position | os*item_id |
| city | tag | page No. | gender*category_id |
| ... | ... | ... | ... |

Sequence Item Features, that is, item class features, can contain many features, but if all the item features are used in the model, the cost is very high, and in practice, category_id item's item_id and category_id 's feature contribute to the characterization of item's information and category_id's effect. Therefore, we only use item_id and category_id' s feature here.

The use of position vector comes from the Transformer model in Attention is all you need published by Google in 2017. Similar to word sequence information in natural language processing, sequence data also exists in the historical behavior sequence data of e-commerce users. position information is supplemented for each click action item in the action sequence. Here, vi is used to represent the item clicked in the behavior sequence, and its corresponding position value pos($v\_i$) = t($v\_t$) -t ($v\_i$), where $v\_i$ represents the item clicked by the user in the behavior sequence, and t($v\_t$) represents the recommended time. This is the time servering requests the model to perform estimation on the midline of the industrial scene. t($v\_i$) represents the timestamp of the user's history of clicking the $v\_i$ item.

### 3.3 Transformer Layer

BST model uses Transformer module, because Transformer can learn the relationship between each item and other items in its behavior sequence, so it can learn the vector representation of each item in more depth. the Transformer structure introduced here includes Self-attention layer, Point-wise Feed-Forward Networks and Stacking the self-attention blocks.

(1) Self-attention layer

For self-attention, the calculation method in Transformer is followed here, and the attention score in the form of dot-product is used, as follows:

$$\text{Attention}(Q, K, V) = \text{softmax}(\frac{QK^T}{\sqrt{d}}) V \quad (3)$$

Q, K and V represent queries, keys and values respectively. In the scenario of e-commerce recommendation, the embedding representation of the item is the input. By linear mapping (that is, the large matrix of the item embedding), they will be transformed into three matrix representations, and then input to the attention layer for calculation, and here maintain the multi-head self-attention in Transformer, that is:

$$S = MH(E) = Concat(head_1, head_2, \cdots, head_h)W^H \quad (4)$$

$$head_i = Attention(EW^Q, EW^K, EW^V) \quad (5)$$

Q, K and V represent queries, keys and values respectively. In the scenario of e-commerce recommendation, the embedding representation of the item is the input. By linear mapping (that is, the large matrix of the item embedding), they will be transformed into three matrix representations, and then input to the attention layer for calculation, and here maintain the multi-head self-attention in Transformer, that is:

$$F = FFN(S) \quad (5)$$

(2) Based on the forward network in Transformer, point-wise granular forward network is used here to further improve the nonlinear expression of the model, namely:

$$S' = LayerNorm\,(S + Dropout(MH(S))) \quad (6)$$

In order to avoid overfitting the model and to be able to learn meaningful features in a hierarchical form, the BST model uses dropout and LeakyReLU algorithms in self-attention and FFN. All outputs of self-attention and FFN are:

$$F = LayerNorm\,(S' + Dropout(LeakyReLU(S'W^{(1)} + b^{(1)})W^{(2)} + b^{(2)})) \quad (7)$$

Where $W^{(1)}$, $b(1)$, $W(2)$, $b(2)$ are all parameters that need to be learned by the model, and LayerNorm is the standard normalization layer.

### 3.4 Model result

To illustrate the effectiveness of BST, we compare it to two models: WDL and DIN. In addition, we create a benchmark, called WDL(+Seq), by integrating the order information into WDL, which aggregates the embeddings of previously clicked items on average. Our framework is built on top of WDL, sequential modeling by adding Transformer, and DIN is proposed by capturing the familiarity between the target item and the previously clicked item with the attention mechanism.

(1) Evaluation indicators

For offline results, we use AUC scores to evaluate the performance of different models. For online A/B testing, we evaluated all models using CTR and average RT. RT is short for Response time and represents the time it takes for a given query to generate recommended results. We use the average RT as a metric to evaluate efficiency in different online production environments.

(2)Result analysis

In order to verify the effect of BST model, offline and online model experiments were conducted on Taobao recommendation service. The baseline model compared is Wide&Deep model and DIN model, and the experimental results are as follows:

| Methods | Offline AUC | Online CTR Gain | Average RT(ms) |
|---|---|---|---|
| WDL | 0.7734 | - | 13 |
| WDL(+Seq) | 0.7846 | +3.03% | 14 |
| DIN | 0.7866 | +4.55% | 16 |
| BST($b=1$) | **0.7894** | +7.57% | 20 |
| BST($b=2$) | 0.7885 | - | - |
| BST($b=3$) | 0.7823 | - | - |

From the experimental comparison of the models, the BST model achieves the best results in both offline and online indicators, which verifies the effectiveness of deep characterization of sequence behavior with Transformer module. At the same time, by adding or not adding sequence behavior features to Wide&Deep, it is verified that sequence behavior itself has significant benefits for model learning users' click interest.

To sum up, the BST model in this experiment realizes better extraction of user behavior sequence characterization information by using Transformer module and position information of behavior sequence. Considering the significant benefits brought by the application of Transformer module in the previous DSIN model, The combined model of recommendation algorithm and Transformer structure is indeed more accurate and rich in the representation of user behavior information, so it is also an important direction for the combination of recommendation field and NLP field in the future.

## 4. CONCLUSION

Based on the comprehensive exploration of the current landscape of large language models (LLMs) and the pivotal role played by Transformer models in e-commerce intelligent recommendation systems, this paper underscores the transformative impact of artificial intelligence on the e-commerce landscape. Through a detailed analysis of Transformer's architecture, including its encoder-decoder structure and multi-head self-attention mechanism, the paper highlights how Transformer models excel in capturing complex relationships within text sequences, making them particularly adept at understanding user behavior and preferences in e-commerce scenarios.

In this study by integrating Transformer models into recommendation algorithms, such as the proposed Behavior Sequence Transformer (BST) model, the paper demonstrates significant improvements in recommendation accuracy and user satisfaction. The BST model leverages Transformer's ability to model sequential information effectively, thereby enhancing the representation of user behavior and item interactions.

Furthermore, the paper emphasizes the broader implications of Transformer models in e-commerce, including their potential to drive innovation in personalized recommendation systems, sentiment analysis, and automated customer service interactions. Additionally, it discusses the ongoing challenges in model generalization, scalability, and privacy protection, highlighting areas for future research and development.

In conclusion, the application of Transformer models in e-commerce intelligent recommendation systems not only signifies a significant technological advancement but also holds immense promise for enhancing user experiences and optimizing business processes. Looking ahead, the fusion of large language models and Transformer architectures is poised to reshape the landscape of artificial intelligence, ushering in a new era of innovation and opportunity across various domains.

# SPIE Proceedings Publications